# A Feature Transfer Enabled Multi-Task Deep Learning Model on Medical Imaging

Fei Gao, Hyunsoo Yoon, Teresa Wu, Xianghua Chu

## Abstract

Object detection, segmentation and classification are three common tasks in medical image analysis. Multi-task deep learning (MTL) tackles these three tasks jointly, which provides several advantages—saving computing time and resources and improving robustness against overfitting. However, existing multi-task deep models start with each task as an individual task and integrate parallelly conducted tasks at the end of the architecture with one cost function. Such architecture fails to take advantage of the combined power of the features from each individual task at an early stage of the training. In this research, we propose a new architecture, FT-MTL-Net, an MTL enabled by feature transferring. Traditional transfer learning deals with the same or similar task from different data sources (a.k.a. domain). The underlying assumption is that the knowledge gained from source domains may help the learning task on the target domain. Our proposed FT-MTL-Net utilizes the different tasks from the same domain. Considering features from the tasks are different views of the domain, the combined feature maps can be well exploited using knowledge from multiple views to enhance the generalizability. To evaluate the validity of the proposed approach, FT-MTL-Net is compared with models from literature including 8 classification models, 4 detection models and 3 segmentation models using a public full field digital mammogram dataset for breast cancer diagnosis. Experimental results show that the proposed FT-MTL-Net outperforms the competing models in classification and detection and has comparable results in segmentation.

## 1. Introduction

During the last decade, precision medicine, an approach that considers individual variability in diagnosis and treatment, has emerged as a novel paradigm for healthcare. One cornerstone for precision medicine is medical imaging. As tremendous resources and manpower have been directed towards research in medical imaging, this domain of study can be broadly divided into three categories: object detection, image segmentation, and imaging-based classification. The aim of object detection is to derive an envelope encircling the object of interest or the center points to locate the objects of interest. Segmentation generates a probability map that quantifies the likelihood of each pixel/voxel being within the region of interest (e.g., tumor). Imaging-based classification primarily identifies the object of interest to be malignant or benign. Most recently, deep learning (Lecun, Bengio, & Hinton, 2015) has gained great success in performing all three tasks (Affonso Carlos, Renato, & Marques, 2015; He, Zhang, Ren, & Sun, 2016; Khatami et al., 2018; Szegedy et al., 2015).

Deep learning owes its success largely to the fact that its models are capable of learning and reproducing an extensive range of parameters from the layers. These parameters are utilized to extract features from images to achieve good performance with respect to the tasks (Litjens et al., 2017). As one of the first deep learning techniques, convolutional neural networks (CNN) (Greenspan, Ginneken, & Summers, 2016) have been extensively investigated. For object detection, CNN-based detectors are trained to find "bounding boxes" on the desired object(s). Example applications are colonic polyps in CT images (Roth et al., 2016), cerebral microbleeds in MRI scans (Dou et al., 2016), and breast and lung cancer in Ultrasound images (Lee & Chen, 2015). For segmentation, successful implementations of CNN have been reported in segmenting brain tumors (Havaei et al., 2017; W. Zhang et al., 2015; Zhao et al., 2018), epithelial tissue in prostatectomy (Bulten, Litjens, Hulsbergen-van de Kaa, & van der Laak, 2018), and joint craniomaxillofacial bone and landmark digitization (J. Zhang et al., 2018). For classification, CNN often takes extracted region of interest (ROI) as the input and the outputs are different class labels on the ROIs. The first application is traced back to 1996 when a 4-layer CNN was employed to classify the ROIs into biopsy-proven masses and normal tissues from mammograms images (Sahiner et al., 1996). Since then different CNNs have been introduced for various medical classification applications including breast lesions (Araujo et al., 2017; Huynh, Li, & Giger, 2016), lung nodules (Shen, Han, Aberle, Bui, & Hsu, 2019), skin lesions (Yap, Yolland, & Tschandl, 2018), and pulmonary peri-fissural nodules (Ciompi et al., 2015), just to name a few. Though commendable classification results have been reported, they are limited only to scenarios where manually labeled tumors (ROIs) are provided.

The research reviewed above focuses on each individual task, namely, detection, segmentation, or classification. Recognizing the inter-dependencies of these tasks, researchers have started to develop pipeline system integrating these tasks where automatic detection and segmentation are often the first steps before classification. For instance, in (Al-Masni et al., 2017), a regional convolutional neural network (R-CNN) is proposed for mass detection, followed by a fully connected CNN-based classifier for "benign versus malignant" prediction. In (Dhungel, Carneiro,

& Bradley, 2017a), a 3-step pipeline for mass detection, segmentation and classification is proposed. Specifically, raw images are fed into a CNN model with the detected candidates being refined through a random forest classifier on hand-crafted features. The refined candidate boxes are then segmented through a Conditional Random Fields (CRF) model (Lafferty, John, Andrew McCallum, 2001) followed by an active contour model (Anne Jorstad, 2014). A mixture model combining CNN model and random forest is trained with bounding boxes extracted from the detection step. The classification results are further tuned through hand-crafted features extracted from both bounding boxes and segmentation outputs from detection. 'User intervention' is introduced where the false positive detections are manually checked and excluded, with the aim of getting an accurate training dataset for the following segmentation and classification tasks. In (Al-antari, Al-masni, Choi, Han, & Kim, 2018), a fully automatic system is designed for detection, segmentation and classification - all deploying deep learning models. You-Only-Look-Once (YOLO) (Redmon, Divvala, Girshick, & Farhadi, 2016) is implemented for mass detection, followed by a full resolution convolutional network (FrCN) for segmentation, and finally a traditional CNN for classification.

Pipeline-based approaches tackle each task one at a time and connect the tasks as a system. Though these approaches perform better in terms of automation or semi-automation and achieve satisfying results on diagnosis, the serial-type pipeline systems come with a set of disadvantages. First, the design and implementation of a deep learning model for each individual task is complicated and time consuming. Large amounts of efforts and computing resources are needed on model design, training, testing and tuning. Second, the relatively limited medical imaging dataset for training could potentially lead to overfitting (Litjens et al., 2017). To address these issues, multi-task learning (MTL) (Caruana, 1997) emerged and has shown great promises in natural language processing (Collobert & Weston, 2008), speech recognition (Deng, Hinton, & Kingsbury., 2013), and computer vision (Girshick, 2015; He, Gkioxari, Dollar, & Girshick, 2017). One advantage of MTL is it saves computational resources by sharing convolutional layers (features maps) amongst separate tasks. MTL also may reduce the risk of overfitting through learning a more generalized feature map for each task (Baxter, 1997; Ruder, 2017). In addition, MTL improves learning efficiency and prediction accuracy for the task-specific models (Caruana, 1997).

Current MTL deep model research is dominated by the direct parameters sharing approach (Ruder, 2017). The models employ "1-m-1" structure. The first "1" is a main shared deep CNN architecture (a.k.a. backbone). The "m" refers to multiple separate subnetworks (a.k.a. head architecture) for different tasks (He et al., 2017). These "m" head architectures share the feature maps from the backbone and make predictions individually. The second "1" is a cost function.

During the training, the parameters from the backbone and the heads are updated simultaneously based on this single cost function in the form of a linear combination of each individual task's cost. Following this "1-m-1" structure, several methods have been proposed for natural image analysis (Redmon et al., 2016; Ren et al., 2017). The success of MTL on natural images is naturally extended to the medical imaging applications. For instance, in (Akselrod-ballin et al., 2016), Faster R-CNN is introduced for detection and classification of mass regions simultaneously. In this architecture, a single ResNet model (He et al., 2016) is implemented to provide mass candidates and feature maps which are shared by the tasks of localization and classification. In (Samala, Chan, Hadjiiski, Helvie, & Cha, 2018), the researchers take mass classification from digital mammograms and digitalized screen-film mammograms as two separate tasks and address these two tasks by a single framework based on the Visual Geometry Group (VGG) model (Noh, Hong, & Han, 2015). Another study (Liu, Zhang, Adeli, & Shen, 2018) focuses on neuroimaging for Alzheimer's disease to diagnose classification and predict clinical scores . In (Feng, Nie, Wang, & Shen, 2018), a multi-task residual fully convolutional network (FCN) is proposed to segment organs (e.g. bladder, prostate and rectum) and estimate the intensities. Here we contend, while "1-m-1" approaches aim to handle multiple tasks from one model, the backbone needs to be carefully designed to include most if not all the features, which must be shared. Moreover, "1-m-1" models fail to consider the potential contributions from the head-features to the tasks, individually and jointly. As medical applications have unique challenges of potential overfitting due to limited training dataset, sharing head-features may help address this issue.

Sharing head-features is an implementation of transfer learning concept. When first proposed, transfer learning was interested in the problems from different data sources. Here the data source is known as the domain. Transfer learning integrates knowledge gained from source domains with the data in target domains to help overcome data shortages in the target domain. The existing transfer learning methods fall into three major categories: instance transfer, parameter transfer, and feature transfer. (Pan & Yang, 2010; Yoon & Li, 2019). Instance transfer reuses data from the source domains to augment the data in the target domains. Although it is intuitive, instance transfer may be questioned for its validity when source and target domains differ greatly. Parameter transfer assumes that closely related tasks should have similar parameters in their respective models and encourages source and target domains to share some model parameters. Yet, it is challenging to appropriately utilize parameters from source domains and tune hyperparameters for the target domain. Feature transfer aims to identify a joint feature map shared by the source and target domains. Because multiple sources and target domains have shared knowledge and representations, features transferred from the source domains may enhance the generalizability of the model with reduced risk of overfitting. However, both parameter and feature transfer face the major obstacle of negative transfer (Pan & Yang, 2010; Yoon & Li, 2019). That is, when domain discrepancy exists, the transferred knowledge may damage instead of helping the predictive power of the models. Fortunately, this research is interested in multiple tasks from the same domain. Considering the feature map from each task is one view of the domain, the domain discrepancy from the cross-domain transferring is not of concern, so the performance of an individual task shall be improved by cross-view feature transferring. Therefore, we propose FT-MTL-Net, an MTL with cross-view feature transferring. It is novel especially for medical application. This is because object detection, segmentation, and classification are three essential and inter-related tasks in medical imaging analysis. Represented joint feature maps from the cross-view feature transfer will take advantage of complementary power of the features from different tasks without having domain discrepancy issue. As a result, the generalizability of target task is enhanced on the medical dataset even with limited samples.

As the initial step to validate the idea of FT-MTL-Net, we decide to explore the feature transferring from the segmentation task to the classification task. This is because 1) the goal of most medical imaging applications is to accurately diagnose/stage the disease - a classification problem; 2) Segmentation and detection are both closely tied to classification, but the features used in segmentation, detection, and classification differ. Specifically, classification and detection require features of low resolution for the abstracted representation (Szegedy et al., 2015; Wu, Zhong, & Liu, 2017), while segmentation needs high resolution features for the pixel/voxel wised prediction (Badrinarayanan, Kendall, & Cipolla, 2017; Shelhamer, Long, & Darrell, 2017). Moreover, given that the segmentation task has already highlighted the candidate areas through the output masks, they can be taken as prior knowledge to guide the feature generation procedure focusing more on highlighted regions to efficiently obtain features representative of the candidate areas. Motivated by these two aspects, our proposed FT-MTL-Net is designed to transfer head features from candidate regions (previously derived from segmentation tasks) to the classification task. Three contributions come out of this novel design. First, to our best knowledge, it is the first fully automatic system for detection, segmentation and classification of tumors in medical imaging which can be trained end-to-end. Second, it enables feature transfer from the segmentation to the classification task. The features from high resolution (transferred from segmentation) and low resolution (existing features) are adopted to help improve the classification accuracy. Third, the features transferred are re-weighted based on the prior knowledge from the segmentation probability map. As a result, information from irrelevant regions is excluded, and the feature map is representative of the tumor regions only. Such design requires less parameters (in this study, 768 parameters) compared to ~2M parameters from Mask-RCNN (He et al., 2017) and thus is computationally efficient.

We evaluate the proposed FT-MTL-Net on INbreast (Moreira et al., 2012), a public full filed digital mammogram (FFDM) dataset. The performance is measured based on five-fold cross validation. For the classification task, this proposed method is compared with 8 methods (4 manual and 4 automated) using the performance metric area under ROC curve (AUC). Experimental results indicate FT-MTL-Net outperforms all eight competing methods with an AUC of 0.92 (± 0.02). For detection task, it outperforms four competing methods with a true positive rate of 0.91 (± 0.05) at an average of 3.67 false positives per image. For the segmentation task, it is compared with 3 existing methods and achieves a comparable result of average dice index of $0.76 \pm 0.03$.

The remainder of the paper is presented as follows. Details about our proposed FT-MTL-Net are presented Section 2. The dataset and processing procedures are discussed in Section 3 followed by the comparison experiments on a public dataset in Section 4. Section 5 draws conclusions with future directions.

## 2. Proposed FT-MTL-Net

The architecture of our proposed FT-MTL-Net is shown in Figure 1. The first part of FT-MTL-Net is the backbone architecture. Similar to Mask-RCNN, the backbone consists of shared convolution layers (Conv layer) for feature map generation and a region proposal network (RPN) (Ren et al., 2017) for candidate region detection. Raw images are fed into the shared convolution layers to generate feature representations for all subsequent tasks (e.g., detection, segmentation and classification). RPN uses bounding boxes with pre-defined sizes to search entire raw images and outputs a set of rectangular candidate regions. Each candidate region is treated as an ROI candidate and has a corresponding area within the feature map to describe it. Feature maps for ROI candidates are resized to be the same through a bilinear interpolation (ROI-align (He et al., 2017)) in order to be fed into the head structures. Following the backbone, three head architectures are proposed to focus on these ROI candidates and make ROI-oriented predictions. Specifically, the detection head is to refine the ROI candidates for an accurate bounding box. The segmentation head generates masks for each ROI candidate. The classification head predicts whether the ROI candidates are benign or malignant.

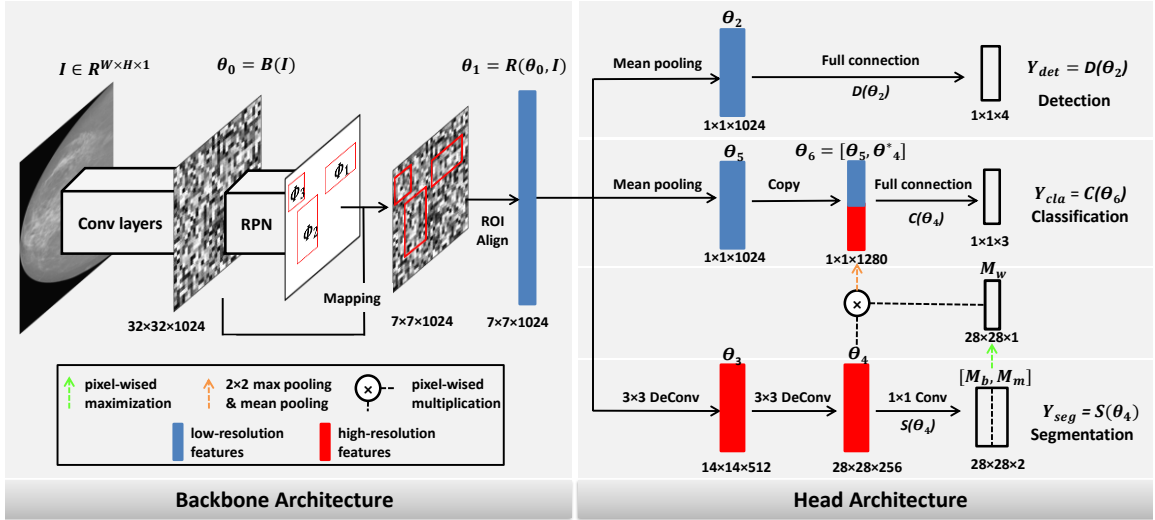

Figure 1 Architecture of proposed FT-MTL-Net

## 2.1 Backbone Architecture

### 2.1.1. Shared Convolution layers for Feature Generation

The first part of the backbone is sharing convolution layers to render feature maps. Note we use 2D images (pixel) in the following discussion for simplicity, and the same methodology applies to 3D images (voxel). Given the gray scale input image $I \in \mathbb{R}^{W \times H \times 1}$, a feature map $\theta_0 = B(I)$ is generated by mapping $B(\cdot)$ conducted by the shared convolution layers. In this research, ResNet (He, Zhang, Ren, & Sun, 2015) is adopted to serve this purpose. ResNet is a well-known deep CNN architecture with the novel design of 'short cut' connection in the building block. Compared to traditional deep-CNNs, this design helps improve the performance in avoiding the problem of gradient vanishing (Drozdzal, Vorontsov, Chartrand, Kadoury, & Pal, 2017; He et al., 2016). Since inception, ResNet has been implemented in various computer vision tasks including medical applications (Fakhry, Zeng, & Ji, 2016; Gao et al., 2018). For the consideration of balance between computation efficiency and accuracy with the limited computation resources, we use ResNet-50. The last fully connected layer originally designed for classification is removed. Note that ResNet has 4 max-pooling layers. Let the original input image be $W \times H \times 1$ (width × height × channel; the following notations of feature map/image size follow this same format, if the channel number equals to 1, it will be omitted), the output of ResNet-50 is a feature map of $w \times h \times 1024$ ($w = W/16$ and $h = H/16$). In this study, the image is $512 \times 512$. As a result, the feature map $\theta_0$ is $32 \times 32 \times 1024$.

### 2.1.2. Region Proposal Network for ROI Proposal Detection

Taking feature map $\theta_0$ from ResNet-50 and raw image $I$ as inputs, RPN (Ren et al., 2017) predicts object bounds and objectness of scores at each position. The objectness score is a probability measure of an object within this specific patch. The outputs are a set of indicators for rectangular candidates (a.k.a. ROI proposals), denoted as $\Phi = \{\Phi_1, \Phi_2, \ldots, \Phi_n\}$. Since the targeting object in the raw image can be at any location with arbitrary sizes, searching the whole raw images for regions of all possible sizes and locations is computationally prohibitive. In RPN, the candidates in $P$ are searched on the feature map using a sliding window. A sliding window runs spatially on the feature map at a pre-defined step size *s*. For each pixel in the center, ROI candidates with pre-defined sizes are generated and mapped back to raw images. For candidate *i*, let $\Phi_i = (a_{iw}, a_{ih}, a_{ix}, a_{iy})$, where $a_{iw}$ denotes the width, $a_{ih}$ denotes the height, and $(a_{ix}, a_{iy})$ denotes the center's coordinate. If $\Phi_i$ has an overlap with the ground truth mask that is greater than a pre-defined threshold, it is taken as a positive ROI candidate. Otherwise, it is negative. Each $\Phi_i$ is represented by a 1-dimension array of features, which is the mean value of each channel on the feature map ($\theta_0$). These features are used to predict the objectness for each $\Phi_i$. After being trained, the RPN will output a set $\Phi$ containing ROI candidates with higher objectness scores than a predefined threshold (e.g., 0.5).

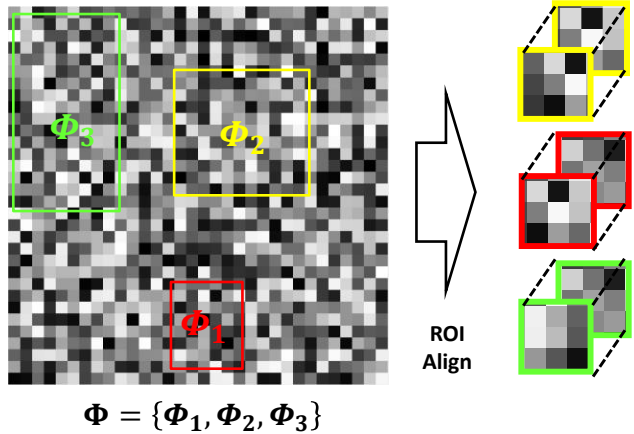

$$\Phi = \{\Phi_1, \Phi_2, \Phi_3\}$$

Figure 2 Illustration for the proposing bounding boxes resized to same size through ROI Align

For $\Phi_i \in \Phi$, the associated bounding boxes on the feature map vary in sizes. Therefore, the candidates are resized to the same size (7×7 in this study) through ROI align layer (He et al., 2017), a linear interpolation procedure. Next, the ROI candidates within $\Phi$ are represented with its associated feature map $\theta_1$ of the same size (as shown in Figure 2), and shared by the head architectures (see section 2.2).

## 2.2 Multi-Task Head Architecture

### 2.2.1. Head Architecture for Detection Task

The detection subnetwork follows the same design in (Ren et al., 2017) where a mean pooling layer is implemented to reduce the feature map resolution to one dimension. It is fully connected to the output layer of bounding box regression. The output value is associated with the corresponding ROI candidate $\Phi_i = (a_{iw}, a_{ih}, a_{ix}, a_{iy})$ before the aforementioned resizing procedure. Let $T = (a_{tw}, a_{th}, a_{tx}, a_{ty})$ be the target candidate, in which $(a_{tx}, a_{ty})$ denotes the predicted center coordinate and $a_{tw}$ and $a_{th}$ denote the predicted width and height, respectively. Assume the targeting outputs for ground truth bounding box is $\Upsilon = (a_{vw}, a_{vh}, a_{vx}, a_{vy})$, where $(a_{vx}, a_{vy})$ denotes the ground truth bounding box's center coordinate, $a_{vw}$ and $a_{vh}$ denote the width and height, respectively. The cost function for regression task is as follows:

$$L_{box}(T, \Upsilon) = Smooth_{L1}(f(T, \Phi_i) - f(\Upsilon, \Phi_i)) \tag{1}$$

where,

$$Smooth_{L1}(x) = \begin{cases} 0.5x^2 & if \ |x| < 1 \\ |x| - 0.5 & otherwise \end{cases} \tag{2}$$

$$f(T, \Phi_i) = \left(\log\left(\frac{a_{tw}}{a_{iw}}\right), \log\left(\frac{a_{th}}{a_{ih}}\right), \frac{a_{tx} - a_{ix}}{a_{iw}}, \frac{a_{ty} - a_{ix}}{a_{ih}}\right) \tag{3}$$

$$f(\Upsilon, \Phi_i) = \left(\log\left(\frac{a_{vw}}{a_{iw}}\right), \log\left(\frac{a_{vh}}{a_{ih}}\right), \frac{a_{vx} - a_{ix}}{a_{iw}}, \frac{a_{vy} - a_{ix}}{a_{ih}}\right) \tag{4}$$

The detection head will refine the sizes and locations of ROI candidates and output the final predictions on the bounding boxes.

### 2.2.2. Head Architecture for Segmentation Task

In the segmentation subnetwork, two deconvolutional layers are introduced to increase the resolution of the feature maps for segmentation and also serve the purpose of deriving task-specific feature maps ($\theta_3$ and $\theta_4$). Following the deconvolutional layers, one $1 \times 1$ convolutional layer is added for the final output. Per-pixel sigmoid function is applied to this final output to obtain two probability maps ($M_b$ and $M_m$). Since the candidate $\Phi_i$ from RPN has 7×7, the resolution is increased by 2x2 (two deconvolution layers) resulting in $M_b$ and $M_m$ sized $\beta \times \gamma$ ($\beta = \gamma = 28$). $M_b$ and $M_m$ describe the probabilities that each pixel is within the benign and malignant tumors independently.

The last feature map ($\theta_4$) before final segmentation output provides high resolution information for each pixel along the 256 channels. The features ($\theta_4$) are different from those from the detection task ($\theta_2$) and classification task ($\theta_5$, discussed in Section 2.2.3). Both $\theta_2$ and $\theta_5$ are abstracted features of lower resolution (He et al., 2017; Ronneberger, Fischer, & Brox, 2015; Shelhamer et al., 2017). We hypothesize that the high-resolution features from the segmentation shall help improve the classification (discussed in Section 2.2.3) greatly, thus they are transferred. Transfering high resolution feature maps to low-resolution feature maps requires some operations. One example is max pooling or average pooling (He et al., 2016; Szegedy et al., 2015) where the maximum or the mean values of the features are derived and transferred. Yet, such an approach treats all features inside and outside ROIs equally. Knowing medical imaging analysis mostly focuses on tumorous areas (such as in this study), we propose a prior knowledge guided feature generation method: feature values representing different regions are re-weighted based on the probability maps. A weight map $M_w$ of the size $\beta \times \gamma$ is generated based on the outputs of segmentation masks $M_b$ and $M_m$:

$$M_{w_{i,j}} = \max\left(M_{b_{i,j}}, M_{m_{i,j}}\right) \text{ for } i \in [1,\beta], j \in [1,\gamma] \quad (5)$$

where $M_{w_{i,j}}$ is combined with the feature map $\theta_4$ (of size $\beta \times \gamma \times \delta$) to generate a prior knowledge guided feature map $\theta_4^* = P(\theta_4, M_w)$ of the same size:

$$\theta_{4_{i,j,k}}^* = \theta_{4_{i,j,k}} \times M_{w_{i,j}} \text{ for } i \in [1,\beta], j \in [1,\gamma], k \in [1,\delta] \quad (6)$$

In order to generate compressed features that can be directly used by the classification task, the resolution of the feature map $\theta_4^*$ is reduced from $28 \times 28$ to $1 \times 1$ through a max pooling layer and a global mean pooling layer (similar procedure as in (Noh et al., 2015)).

For the cost function of segmentation, assume the output prediction map is $s^{\beta \times \gamma}$ of resolution $\beta \times \gamma$ and the cost function for segmentation is average cross entropy over all the pixels within the s and ground truth mask $m^{M \times N}$, which can be calculated as follows:

$$L_{mask}(s,m) = \left(\frac{1}{\beta \times \gamma}\right) \sum_{i=1}^{\beta} \sum_{j=1}^{\gamma} CrossEntropy(s_{ij}, m_{ij}) \quad (7)$$

in which

$$CrossEntropy(y^*, y) = -y\log(y^*) - (1-y)\log(1-y^*) \tag{8}$$

The segmentation head outputs two individual probability maps measuring likelihood of each pixel being within benign and malignant tumor respectively. Following the same setting as Mask-RCNN (He et al., 2017) to solve the overlapping issue of different types tumors, a final mask is selected based on the output of the classification task.

### 2.2.3. Head Architecture for Classification Task

In the classification subnetwork, the feature map $\theta_6$ for the final classification layer is of size $1 \times 1 \times 1280$. Among these 1280 feature channels, 1024 are obtained from a shared feature map $\theta_1$ provided by the backbone through a global mean pooling. The remaining 256 channels come from $\theta_4'$, which are used as an addition of pixel wised information. The feature channels from two sources are combined together and fully connected to the final classification layer with 3 outputs (background, benign and malignant), and a corresponding probability array $P = (p_0, p_1, p_2)$ is computed by a softmax activation function (Krizhevsky, Sutskever, & Hinton, 2012) over the 3 outputs. The cost function for the classification task is the log loss function for its corresponding class $u$:

$$L_{cls}(p, u) = -log(p_u) \tag{9}$$

The classification head outputs final predictions for the detected ROIs to be: background, benign or malignant. The ROIs with high probabilities of being benign or malignant tumors are investigated for final prediction using "malignant-veto" logic described in Section 2.3.

### 2.3 Model Training and Inference

Table 1 Detailed steps of the training procedure

Step 1. Initialize the ResNet-50 with the weights trained using natural images from the dataset of ImageNet, which is made available online by the developers of ResNet (He et al., 2016).

Step 2. Initialize the weights of all other layers through a normal distribution with mean = 0 and standard deviation = 0.05.

Step 3. Fine-tuned end-to-end for the region candidate task using cost function $L_{prop}$.

Step 4. Keep the weights within shared layers and RPN layer fixed, tune the weights within subnetworks alone with cost function $L_{uni}$.

Step 5. Keep tuning the weights within shared layers and subnetworks together with cost function $L_{uni}$.

In the training procedure, all three tasks are trained simultaneously with one combined loss function:

$$L_{uni} = \lambda_1 L_{cls} + \lambda_2 L_{box} + \lambda_3 L_{mask} \qquad (10)$$

where $\lambda_1, \lambda_2$, and $\lambda_3$ are weights for each individual cost function. In this study, $\lambda_1, \lambda_2$, and $\lambda_3$ are all set to be 1. A 5-step training procedure (see Table 2) following the same logic in (Ren et al., 2017) is adopted. Once the training process is complete, the model is ready to make inferences for testing images.

There are two major differences between the inference workflow and the training procedure. The first difference is sequential execution vs. parallel training. That is, in inference, it follows (Step 1) ROI candidates are obtained from the backbone; (Step 2) the detection task is conducted to provide accurate bounding box predictions; (Step 3) the segmentation task is triggered to generate mask predictions and features based on the bounding boxes; (Step 4) features from segmentation are transferred and joined for classification. The second difference is an added "malignant-veto" logic motivated by the medical practices in the inference workflow. As expected, each medical case often may have multiple bounding boxes and thus ROIs to be investigated. We define the "malignant-veto" logic as: if one bounding box is predicted as malignant, this mass will be predicted as malignant with a score equaling the maximum score among all these boxes indicating malignancy; if none of the bounding boxes indicates malignancy, it gets a malignancy score $[1 - Sb_{max}]$, where $Sb_{max}$ is the maximum score among all the bounding boxes assigned with a benign score.

## 3. Experiment and Results

### 3.1 Dataset

The dataset used in this study is obtained from INbreast, an online accessible full-field digital mammographic (FFDM) database (Moreira et al., 2012). INbreast was established by the researchers from the Breast Center in CHJKS, Porto, under the permission of both the Hospital's Ethics Committee and the National Committee of Data Protection. The FFDM images were acquired from the MammoNovation Siemens system with pixel size of 70 mm (microns), and 14-bit contrast resolution. The resolution of each image is 2560 × 3328. For each subject, both CC and MLO view were available. For each image, the annotations of region of interests (ROIs) (ground truth masks) were made by a specialist in the field and were validated by a second specialist. The ROI masks were also made available through the attached xml file.

In this research, 108 subjects with labeled masses are selected. Each mass is assigned with a Breast Imaging Reporting and Data System (BI-RADS) (Eberl, Fox, Edge, Carter, & Mahoney, 2015) score ranging from 2 to 6. Following the same definition in (Dhungel et al., 2017a), the masses with BI-RADS score=2, 3 are treated as benign and the remaining cases (BI-RADS=4, 5, 6) are labeled as malignant. There are 37 benign subjects and 71 malignant subjects.

## 3.2 Data pre-processing

For cases with multiple masses in one image, each individual mass and its corresponding bounding mask is extracted and saved as a new data sample. As a result, the total number of cases in the dataset increases to 115 (41 benign vs. 74 malignant). For each mass, a bounding box is computed as the minimal rectangle in the image that contains the whole mass. In the second step, for each breast image, a rectangle that contains the entire breast is obtained, and the region outside of this bounding box is excluded. This step is to exclude background region in each image and reduce search space and computational burden during training process.

Five-fold cross validation is adopted and data augmentation is implemented to enrich the training dataset. Specifically, within each fold, the training dataset (80%) is augmented by randomly selecting 2 to 5 options from the operations including rotating, flipping, zooming in/out, cropping, contrasting enhancement and Gaussian smoothing. The image, mask and bounding box will go through the same procedure. Considering the balance of benign cases vs. malignant case, each benign sample is augmented 150 times, and each malignant sample is augmented 75 times. The final training dataset has 8,700 images (4,440 benign vs 4,260 malignant).

## 3.3 Experimental Setup

The experiments are conducted on a windows desktop with 32G RAM and an Intel 16-core

CPU. The model is trained using one single NVIDIA Titan XP GPU with 12G memory. Both the data processing procedure and the architecture are developed with Python and deep learning libraries (e.g., Keras and tensorflow). The whole architecture is built upon the MASK RCNN package downloaded through the open source website GitHub (https://github.com/matterport/Mask_ RCNN). Details of tuned parameters are: (1) training iterations for the 4 training steps are set to be 10; (2) the learning rate for each step is set to be 0.005 with a momentum equal to 0.9; (3) the training batch size is set to be 8 to satisfy the GPU memory; (4) other parameters are set with default values provided by Keras or the downloaded Mask RCNN package.

## 3.4 Experimental Results

FT-MTL is designed for three inter-related tasks in medical applications: classification, object detection and segmentation. High resolution features from the segmentation task are transferred to the classification task for improved performance. In the comparison study, we decide to compare the proposed FT-MTL with methods in classification, object detection and segmentation, respectively. These include some methods that only focus on one of the three tasks, e.g., classification, as well as methods handling multiple tasks. To the best of our knowledge, Mask-RCNN (He et al., 2017) may be the only method that addresses all three tasks jointly for medical applications. We include Mask-RCNN in the comparison on all three tasks with the competing methods respectively. In addition, detailed comparison analysis between FT-MTL and Mask-RCNN is provided.

### 3.4.1. Classification Task

A response operating characteristic (ROC) curve is commonly used to evaluate the classification performance, especially in medical imaging applications. ROC is a function of true positive rate (TPR) with respect to 1- false positive rate (1-FPR). The area under the ROC curve (AUC) is used as a metric to evaluate the classification power of a model. Table 2 summarizes the comparison results. The first three methods take manually delineated ROIs from domain experts as inputs and focus on the classification task only. The AUC ranges from 0.86 to 0.91. The following four pipelined systems are automated systems taking the whole images detecting the objects, and classifying them . Here we take the classification results for comparison, and the AUC ranges from 0.76 to 0.86. It is not surprising the AUC performances from the pipelined system are not as good as that from the one-task approaches as the later heavily involves the domain experts to provide the accurate segmentations. However, the delineation of the ROIs by experts is time consuming and may not always be available. In looking at the multi-task category, we observe the approaches in the category outperforms most one-task and pipelined systems. Though Mask-RCNN has an AUC of 0.89, lower than that from Random Forest on CNN (0.91), Mask-RCNN has much smaller standard deviation, 0.02 compared to 0.12 from the Random Forest, an indicator of the robustness of the model.

Table 3 Comparison between our proposed model and eight competing methods on Mass classification with INBreast Dataset

| Methods | Configuration | AUC |
| --- | --- | --- |
| Transfer learning from deep CNNs + ensembled classifiers (Huynh et al., 2016) | one task | $0.86 \pm 0.01$ |
| Lib SVM (Diz, Marreiros, & Freitas, 2016) | one task | 0.90 |
| Random Forest on CNN with pre-training (Dhungel et al., 2017a) | one task | $0.91 \pm 0.12$ |
| Random Forest on CNN with pre-training (Dhungel et al., 2017a) | pipelined system | $0.76 \pm 0.23$ |
| Multi-view Residual Network (Dhungel, Carneiro, & Bradley, 2017b) | pipelined system | $0.80 \pm 0.04$ |
| Deep learning through unregistered views (Carneiro, Nascimento, & Bradley, 2017) | pipelined system | $0.78 \pm 0.09$ |
| Pre-trained CNNs + multiple instance learning (Zhu, Lou, Vang, & Xie, 2017) | pipelined system | $0.86 \pm 0.03$ |
| Mask-RCNN (He et al., 2017) | multi-task | $0.89 \pm 0.02$ |
| **Proposed FT-MTL** | **multi-task** | **$0.92 \pm 0.01$** |

In comparing our proposed FT-MTL with Mask-RCNN (see Figure 3), the ROC curve from FT-MTL in general dominates that from Mask-RCNN. FT-MTL has AUC $0.92 \pm 0.01$ compared to Mask-RCNN with $0.89 \pm 0.02$. A pair t-test gives $p<0.01$ indicating FT-MTL significantly outperforms Mask-RCNN on AUC value.

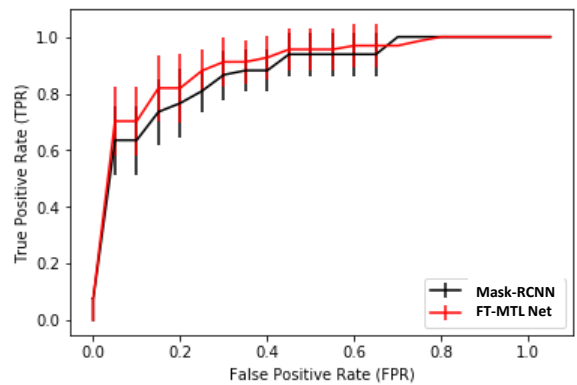

Figure 3 ROC curves for Mask-RCNN and our proposed model on test dataset for detected mass **(vertical**

line denotes 2×TPR std across 5 folds)

From this comparison, we have two conclusions drawn: (1) a multi-task learning approach outperforms traditional one-task and integrated pipeline approaches, indicating the joint advantages of multiple tasks; (2) features transferred from the segmentation head to the classification head will significantly improve the classification performance.

Please note as the first attempt into MTL, our current design of FT-MTL only transfers the segmentation features into the classification. As one MTL approach in general, we are still interested in exploring the performance of the detection and segmentation tasks with respect to competing methods. This is discussed in the following two sections.

### 3.4.2. Detection Tasks

For the detection experiment, we first present the comparison results in mean true positive rate (TPR) across 5 folds and false positive rates per image (FPI) (see Table 3). Since the literature reported the TPRs under different FPIs, for a comprehensive and fair comparison, we derive two sets of TPRs under different FPI settings: FPI = 3.67 and/or 5. Standard deviation across 5 folds is reported. As seen from Table 3, multi-task learning approaches (Mask-RCNN and FT-MTL-Net) have comparable detecting power as traditional one-task detection models and pipelined systems. It should be noted that the Multi-view Residual Network (Dhungel et al., 2017b) achieves the best performance (0.96±0.03@0.8). This is because after the detection module, a specifically designed cluster method is implemented to remove overlapping for both true positives and false positives. It is our intention to further improve the detection performance by adopting some new postprocessing methods such as those in Dhungel et al., (2017b).

Table 4 Comparison between our proposed model and other competing methods on detection with INBreast Dataset

| Methods | Configuration | TPR@FPI |
|---|---|---|
| Adaptive thresholding + machine learning (Kozegar, Soryani, Minaei, & Domingues, 2013) | one task | 0.84@3.67 |
| Cascaded Deep Learning +Random Forests (Dhungel, Carneiro, & Bradley, 2015) | one task | 0.78@3.67 |
| Random Forest on CNN with pre-training (Dhungel et al., 2017a) | pipelined system | 0.87@5 |
| Multi-view Residual Network (Dhungel et al., 2017b) | pipelined system | 0.96±0.03@0.8 |
| Deep learning through unregistered views (Carneiro et al., 2017) | pipelined system | N.A. |
| Pre-trained CNNs + multiple instance learning (Zhu et al., 2017) | pipelined system | N.A. |
| Mask-RCNN (He et al., 2017) | multi-task | 0.85 ±0.07@3.67<br>0.85 ± 0.07@5 |
| **Proposed FT-MTL-Net** | multi-task | **0.91 ±0.05 @3.67**<br>**0.91 ± 0.05@5** |

Next, we compare FT-MTL-Net with Mask-RCNN. Here we use the free response operating characteristic (FROC) curve to present its performance. It is a function of true positive rate (TPR) with respect to false positive rate per image (FPI). Following the same standard in (Dhungel et al., 2017a), we define: if the intersection of union (IoU) between predicted bounding boxes and ground truth is greater than 0.2, this bounding box is regarded as true positive, otherwise, it will be regarded as false negative. From 错误!未找到引用源。a, we observe that FT-MTL-Net achieves a TPR of 0.91 with standard deviation of 0.05 (TPR = 0.91 ± 0.05 )at FPI = 3.67 on the testing dataset. In fact, this TPR (0.91) tends to be stable for FPIs that are greater than 1.5. The Mask-RCNN obtains a TPR = 0.85 ± 0.07 at FPI = 3.67. A t-test is conducted on the TPR values obtained among the 5 folds for Mask-RCNN and our proposed model. With a p-value < 0.05, we conclude FT-MTL-Net outperforms Mask-RCNN. One may be surprised to observe such performance as our FT-MTL-Net indeed takes the same architecture as that in (Ren et al., 2017) for the detection task. This may be explained as following: in the testing stage, each detected bounding box uses the probabilities (background vs. benign tumor vs. malignant tumor) from the classification task as its objectness score. FT-MTL-Net has a classification head architecture with enhanced capability which is not only better at differentiating benign tumors from malignant tumors, but also better at classifying tumors from background regions. This capability in turn helps improve the detection task indirectly. To measure the robustness of the detection results on different IoU thresholds, the average precision curve is shown in 错误!未找到引用源。. It is a function of true positive rate against the different IoUs. It is noted for values where IoU <= 0.4, the TPR remains stable and consistently is above 0.9. The TPR starts to decrease if IoU is greater than 0.4. As a result, we set IoU = 0.4 as our threshold to define whether a mass is detected by the predicted bounding box for the following two tasks. The performances for segmentation and classification are evaluated only on the detected mass which takes an average of 95% of the testing dataset according to the curve. In integrated systems such as in (Dhungel et al., 2017a), similar approaches are implemented by manually excluding all false positives.

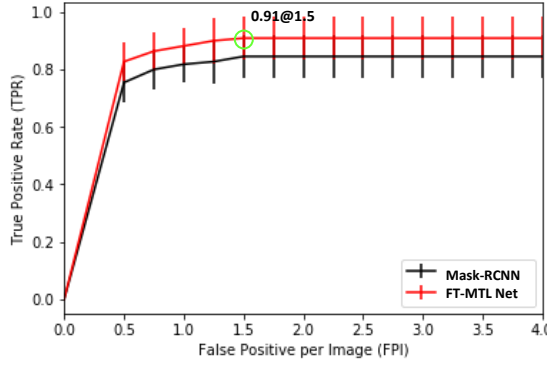 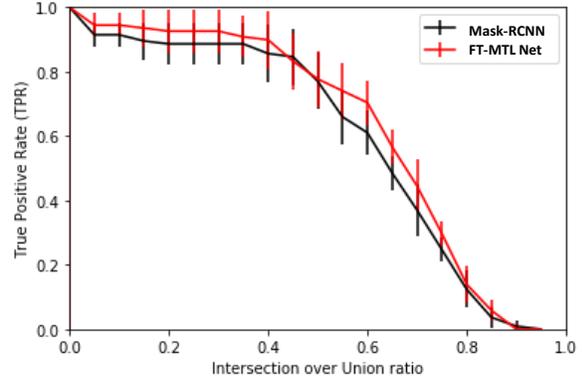

**Figure 4a FROC curve (IoU > 0.2, vertical line denotes 2×TPR std across 5 folds)**

**Figure 5b Average precision for detection on the testing dataset (vertical line denotes 2×TPR std across 5 folds)**

### *3.4.3. Segmentation Tasks*

The segmentation performance is quantified with Dice similarity index (Dice, 1945). Let A be the predicted mask, and B be the ground truth mask,

$$Dice(A,B) = \frac{2(A \cap B)}{A \cup B} \tag{10}$$

Where

$A \cap B$ counts the number of pixels that are labeled with 1s in both mask A and B.

$A \cup B$ counts the number of pixels that are labeled with 1s either in mask A or B.

We compare FT-MTL-Net with Mask-RCNN, 1 one-task method and the same four pipelined systems. From Table 5, we observe these two MTL models underperform the other competing method to a certain degree. The reason may be that, in (Dhungel et al., 2017a) and (Al-antari et al., 2018), the input training images are outputs from a former detection procedure, there is a 'manual intervention' procedure that will exclude all the false positive detections and this helps improve the performance of segmentation results. The MTL models are fully automatic model without any user intervention. The segmentation network is trained with both true positive and false positive detections from the RPN, and the false positive detections have a negative influence on segmentation results. Another reason may come from an architecture aspect: the feature maps used for segmentation are highly reduced in spatial resolution compared with the original masks. Before the segmentation network, 4 max-pooling layers are implemented within the shared convolutional layers, in which important pixel information for segmentation are lost (Chen et al., 2017). Such lost information is difficult (if not impossible at all) to retrieve through the subsequent layers. With limited pixel information, the segmentation network may suffer from low accuracy. Noting this, our plan for the next steps is to improve FT-MTL with a focus on segmentation improvement. For example, we may add a connecting path from high-resolution features to enrich feature sets as those in Unet (Gao et al., 2019; Ronneberger et al., 2015) and SegNet (Badrinarayanan et al., 2017).

Table 5 Comparison between our proposed model and other competing methods on segmentation with INBreast Dataset

| Methods | Configuration | DICE index |
|---|---|---|
| FrCNN (Al-antari et al., 2018) | one task | 92.67 |
| Random Forest on CNN with pre-training (Dhungel et al., 2017a) | pipelined system | $0.85 \pm 0.02$ |
| Multi-view Residual Network (Dhungel et al., 2017b) | pipelined system | N.A. |
| Deep learning through unregistered views (Carneiro et al., 2017) | pipelined system | N.A. |
| Pre-trained CNNs + multiple instance learning (Zhu et al., 2017) | pipelined system | N.A. |
| Mask-RCNN (He et al., 2017) | multi-task | $0.79 \pm 0.02$ |
| **Proposed FT-MTL-Net** | multi-task | **$0.76 \pm 0.03$** |

### *3.4.4. Illustration*

To demonstrate the functions of FT-MTL-Net, we select predicting results for two cases and their corresponding outputs after different steps and include them in Figure 6. As shown, each raw image is fed into the trained model. After the backbone architecture, several candidates (marked with yellow dashed bounding box) of pre-defined size and with objectness score (O score) greater than 0.5 are detected (the above case has two candidates and the bottom case has only one). These candidates are resized to the same size and fed into the head architecture. Through the head architecture, each candidate's bounding box (dashed bounding boxes) will be refined by the detection task; the mask (solid contour region) will be predicted through the segmentation task; the classification task will assign each candidate a probability of being malignant or benign (M score/B score). These predicted results will be finalized through the "malignant-veto" logic introduced above to reduce the overlapping detections. In this figure, we can conclude that our proposed method can 1) accurately identify suspicious regions within breast images 2) make precise predictions on the suspicious regions' categories and 3) outputs segmentation masks with reasonable accuracy.

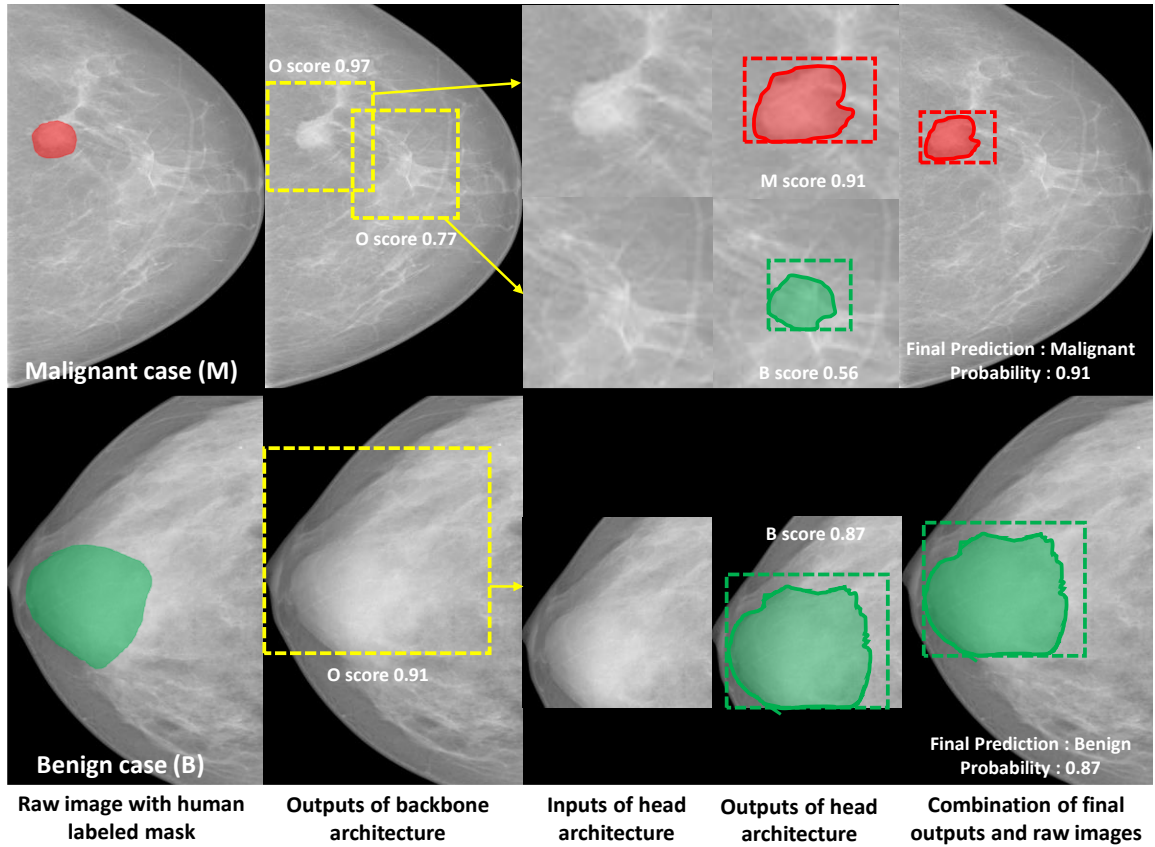

Figure 6 Examples of two cases (malignant case on top and benign case on bottom) and their corresponding outputs from different steps.

## 4. Conclusion and Discussion

Most medical image analysis applications are related to one or more tasks in object detection, segmentation and classification. Multi-task deep learning thus becomes a viable solution to address these tasks jointly as it provides the advantages of both multi-task learning and deep neural networks and enables fully automatic image analysis. In this research, we propose a new multi-task deep learning architecture with a focus on the classification task. Cross-view feature transfer based on the same domain knowledge is implemented. To the best of our knowledge, our proposed FT-MTL may be the first fully automatic system which addresses detection, segmentation and classification of tumors in medical imaging and can simultaneously be trained end-to-end. The feature transfer is implemented between the different views from the same domain and thus is negative transfer free. In addition, the features transferred are re-weighted based on the targeted ROIs resulting in much adding a much smaller number of parameters to the model. The immediate advantage of this is the model does not have a significant computational burden added even with all three tasks being tackled together. The comparison experiments indicate the promise of this cross-view feature transferred enabled MTL. While promising, as we mentioned in the discussion related to the detection and segmentation tasks, we plan to explore the features transferred across all three tasks to improve the performance of all three tasks together. This can be further validated in other clinical applications (e.g., brain tumor) with different imaging modalities (e.g., MR, PET).